# Bald Eagle Search Algorithm for High Precision Inverse Kinematics of Hyper-Redundant 9-DOF Robot

Vineeth P.,[1] Guru Nanma P.,[2] V. Sankar,[1] B. Sachin Kumar[1*]

[1]Department of Mechanical Engineering, B.M.S. College of Engineering, Bengaluru 560019, India.

[2]Department of Computer Science and Engineering, B.M.S. College of Engineering, Bengaluru 560019, India.

*Corresponding author email id: sachinkumarb.mech@bmsce.ac.in

*Corresponding author ORCiD: 0000-0002-8042-0894

# ABSTRACT

Robots in 3D spaces with more than six degrees of freedom are redundant. A redundant robot allows multiple configurations of the robot for the given target point in the dexterous workspace. The presence of multiple solutions helps in resolving constraints in workspace such as object avoidance and energy minimization during trajectory planning. Inverse kinematics solutions of such redundant robotics are intricate. The present study involves comparison of different metaheuristic optimization algorithms (MOA), which have a positional error, and identify a MOA for high precision of positioning of the end effector of the robot. This study applies recent MOA for the inverse kinematics of hyper redundant nine degrees of freedom (DOF) robot arm by using forward kinematics of the Denavit-Hartenberg (DH) parameters and compares the performance of these algorithms. The comparative study shows Bald Eagle Search (BES) algorithm has better performance over other metaheuristic algorithms. BES algorithm outperforms the other MOA in achieving the desired position with very high precision and least positional error for a 9-DOF robot arm.

# 1. INTRODUCTION

Inverse kinematics (IK) is the technique used to find the joint variables of a robotic arm, given the final position of the end effector of the robot arm that needed to be achieved or reached. Standard IK solutions involve the usage of Jacobian matrices,, and the solution of multivariable differential equations that often result in multiple, and complex solutions. In the case of hyper-redundant robots, the number of solutions to a standard inverse kinematic equation will be infinite. Therefore, it is preferred to use metaheuristic optimization algorithms (MOA) for the inverse kinematic problem to attain high precision position.

MOA is used to solve a maximization or minimization problem. MOA is also used for producing effective solutions to nondeterministic polynomial type problems in different fields. MOA searches the solution space for the optimum solution for a given objective function through an iterative process. For a hyper redundant robot arm, infinite solutions are present to reach a particular position. This results in a complex process in determining the joint angle of the hyper redundant robot arm in a Cartesian coordinate through conventional methods such as geometric, and algebraic methods. The inverse kinematic problem for a 9-DOF robot consists of nine different joint variables for a given desired position of the end effector of robotic arm. The goal of MOA is to return the values of nine joint variables such that there is minimum error between the calculated position reached by the end effector, and the desired end effector position. This is achieved by minimizing the fitness function in MOA. MOA runs multiple iterations, where it assumes a different joint variable value to reach a desired end effector position. The values obtained are then substituted in the objective function to find the value of the fitness function. The particle with the least value of fitness function is then selected, and used as a reference for the other particles to find the global minima. The comparative study of different metaheuristic algorithms helps in the selection of high precision algorithm for achieving the desired configuration of the robot.[1]

The four heuristic methods, Genetic Algorithm (GA), Particle Swarm Optimization (PSO) algorithm, Quantum Particle Swarm Optimization (QPSO) algorithm, and Gravitational Search Algorithm (GSA), were used for the IK solution of a 4-DOF real robotics manipulator, and have compared the results obtained, and found that QPSO performed better than the other algorithms in terms of positional error.[2] Simulated Annealing was also applied to solve IK problem of 2-DOF planar robot, and reported that IK problems could be solved even at points of singularity.[3] PSO was applied to solve inverse kinematics problem for a 7-DOF robot, and simulation studies was conducted to validate its effectiveness.[4–7] Additionally, PSO was also used to solve the 6-DOF of inverse kinematics problems, and the results of this study demonstrated the application of MOA to inverse kinematics was efficient.[8] A heuristic method was developed based on firefly algorithm for 3-DOF robot, and reported a positional error of $10^{-8}$ units.[9] An attempt was also made to compare GA, and Hybrid Genetic Algorithm (HGA) for solving inverse kinematics of PUMA (RRR) robot, and it was found that HGA had better accuracy, and convergence rate than GA.[10–13]

In a comparison study for a 7-DOF robot, QPSO performed better than Artificial Bee Colony (ABC), PSO, and Firefly algorithms. QPSO was assigned a particle size of 150 (500 iterations) while ABC, Firefly, and PSO were assigned a particle size of 100, 50, and 300, respectively. QPSO had the least mean squared error (MSE) of $6.9 \times 10^{-13}$ cm followed by ABC, Firefly, and PSO with a MSE of $10^{-6}$ cm, $10^{-5}$ cm, and $10^{-4}$ cm, correspondingly.[14,15] Bald Eagle Search (BES) was chosen as it combines both swarm intelligence, and evolutionary algorithm to solve optimization problems.[16] Physics-based algorithms such as Electromagnetic Field Optimization (EFO), Multiverse Optimization (MVO), and Nuclear Reaction Optimization (NRO) was used as they showed high accuracy, and fast computational time for benchmark problems.[17–19] To the best of our knowledge, it was evident from the literature that these MOAs were not applied to solve IK problem. Further, evolutionary-based algorithm, Coral reef Algorithm (CRO), was applied as it demonstrated exactness for benchmark problems.[20]

## 2. METHODOLOGY

MOA requires parameters such as the number of particles, the number of iterations, constraints of the robot joints, and the boundaries of the workspace. These parameters are varied according to the construction or design of the robot, the shape of the workspace, and the presence of any obstacles in the workspace. These parameters can also determine the accuracy of the solution provided by the MOA. The parameters used in the MOA are: number of variables is nine, number of particles if twenty and number of iterations is five hundred. All the algorithms have been coded on Python (3.8.0), and executed on a computer with the following specification: Model is HP Intel(R) Core (TM) i3-7100U CPU @ 2.40 GHz, and Installed RAM is 4.00 GB.

The number of variables is assigned as nine as the robot being used is a 9-DOF robot, and has nine revolute joints, and the configuration of the robot is visualized in RoboAnalyzer (V7.5 June 23rd 2019).[21] The number of particles, and number of iterations were determined through multiple iterations of testing, this was done in order to minimize the program runtime, and the positional error.

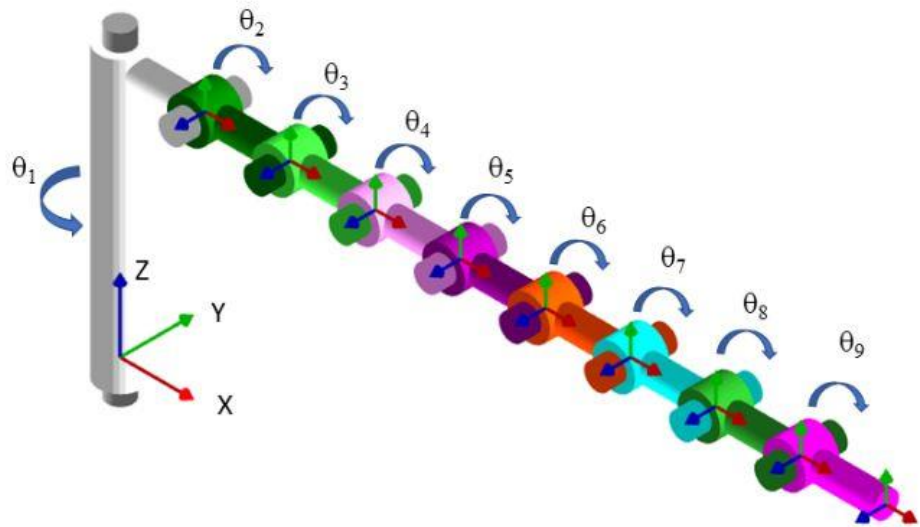

**Figure 1:** Configuration of Robot.

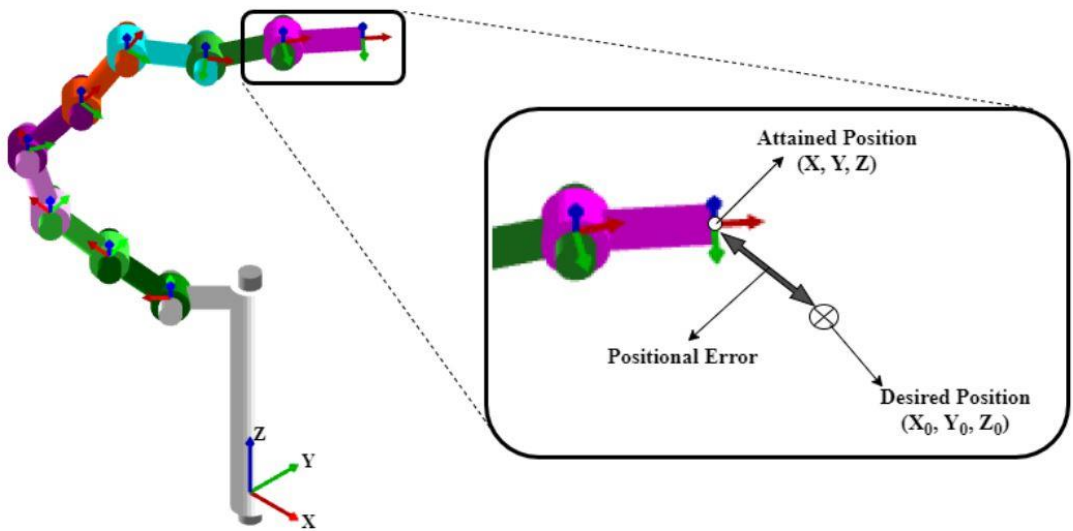

**Figure 2:** Positional error attained from MOA for a desired position.

**Table 1:** Denavit-Hartenberg (DH) parameters.

| Joint number | a (cm) | α (°) | d (cm) | θ (°) |
|---|---|---|---|---|
| 1 | 0 | 90 | 3 | (0, 360) |
| 2 | 1 | 0 | 0 | (-90, 90) |
| 3 | 1 | 0 | 0 | (-90, 90) |
| 4 | 1 | 0 | 0 | (-90, 90) |
| 5 | 1 | 0 | 0 | (-90, 90) |
| 6 | 1 | 0 | 0 | (-90, 90) |
| 7 | 1 | 0 | 0 | (-90, 90) |
| 8 | 1 | 0 | 0 | (-90, 90) |
| 9 | 1 | 0 | 0 | (-90, 90) |

Figure 1 shows the robot used for the purpose of this study, which is a 9-DOF hyper-redundant robot arm. The robot arm anatomy consists of nine serial rotary joints. Figure 2 shows the positional error which is attained when MOA is applied to the IK problem. The Denavit-Hartenberg (DH) Parameters of the robot is as shown in Table 1, where, the angle about common normal is $\alpha$ (*i.e.,* previous link z axis, and the present link z axis), $d$ is offset about previous z-axis, and common normal, $a$ is length of common normal, $\theta$ is angle about previous z-axis.[22]

The Transformation Matrix used for forward kinematics is obtained by,

$$^{n-1}_{n}T = \begin{pmatrix} \cos\theta_n & -\sin\theta_n \cos\alpha_n & \sin\theta_n \sin\alpha_n & r_n \cos\theta_n \\ \sin\theta_n & \cos\theta_n \cos\alpha_n & -\cos\theta_n \sin\alpha_n & r_n \sin\theta_n \\ 0 & \sin\alpha_n & \cos\alpha_n & d_n \\ 0 & 0 & 0 & 1 \end{pmatrix}. \quad (1)$$

The transformation matrix for the nine joints was obtained using the above equation (1), and the DH Parameters of the respective joint. Further the overall transformation matrix of the robot arm is obtained by,

$$T = T1 \times T2 \times T3 \times T4 \times T5 \times T6 \times T7 \times T8 \times T9, \quad (2)$$

where, *T* is a 4 × 4 matrix, and the (*x, y, z*) coordinates of the end effector of the robot is given by the 1st, 2nd, and 3rd elements, respectively, of the 4th column of the matrix, $T$. The equation for $(x, y, z)$ was obtained by equation (1), and (2) are used in the fitness function for MOA, which is given by,

$$f = \sqrt{[(x_0 - x)^2 + (y_0 - y)^2 + (z_0 - z)^2]}. \tag{3}$$

where, $(x_0, y_0, z_0)$ is the Cartesian coordinate of the desired position, and , $(x, y, z)$ is the position obtained from the MOA whose fitness is being evaluated using Euclidean distance equation between two points.

## 3. OPTIMIZING ALGORITHMS

### 3.1 Particle Swarm Optimization

PSO was inspired by the concept of swarm intelligence shown by flocks of migratory birds, and schools of fishes. It is one of the most well researched, and widely used MOA. In this technique, a few candidate solutions are set, and the optimal solution is found through communication, and learning from the performance of each particle.[23]

At first, 20 particles are defined at random points across the robot arms solution space, and run through the MOA to check the fitness function. In each iteration, a swarm best is defined for every swarm, and a global best is defined for all the particles. The position of each particle $P(i)$, the swarm best $P_s(i)$, and the global best $P_g(i)$ are recorded. A vector is defined for every particle as a sum of the original velocity $V(i)$ of the particle, the relative direction of the swarm best $P_g(i) - P(i)$, and the relative position of the global best $P_g(i) - P(i)$. This vector represents the updated velocity of that particular particle $P(i)$. Subsequently, this method is followed for each particle in the iteration, and 500 iterations are executed until all the particles reach the global optimal solution. The computational parameters used for PSO are shown in Table 2. The mathematical model of PSO is given by,

$$\vec{V}(i+1) = \vec{V}(i) + [\vec{P}s(i) - \vec{P}(i)] + [\vec{P}g(i) - \vec{P}(i)] . \tag{4}$$

### 3.2 Coral Reef Optimization

CRO is a modified swarm-based evolutionary MOA that emulates coral reefs. CRO uses methods of sexual, and asexual reproduction to determine which coral (position) is healthy (optimum) when using the health function (fitness function).

CRO algorithm is executed into two phases to obtain the optimal solution. In the first phase the reef is initialized, where a $1 \times 20$ matrix is pre-defined in which a particular number of elements are filled as initial position, whereas others are left empty. In each iteration, the health of the corals is tested against the health function, in this study being the positional error. The healthier corals are allowed to occupy the matrix while the unhealthier ones are released. The rate between releasing, and occupying of the corals in the reef is 0.4. The reef will continue to progress towards the optimum solution as long as the healthy corals stay, and the unhealthy corals are released.

Later, the second phase of CRO, a reef is formed using sexual, and asexual reproduction to determine the optimum solution of the objective function. The corals in the reef were divided into two parts. The first part is used to carryout sexual reproduction while the second part is used for asexual reproduction. Sexual reproduction is further divided into internal, and external sexual reproduction. In external sexual reproduction (*i.e.,* Broadcast Spawning), the corals are made to form pairs of parent corals which crossover to form larvae (*i.e.,* new positions). In case of internal sexual reproduction (*i.e.,* Brooding), each coral is mutated to form new larvae. All the newly formed larvae now attempt to occupy the reef. It has to be noted that if the larva attempts to settle in an uncontested spot, it is allowed to remain in that spot until the next iteration. Otherwise, if a spot in the reef is contested by more than one larva, the healthiest larva is given the spot. Each larva is given 500 iterations to compete for a spot in the reef before it is released.

In an asexual reproduction, all of the existing corals in the reef are sorted according to their health, and the healthiest of these corals are duplicated. The duplicated corals are sent to a different part of the reef to settle as larvae. At the end of each reproductive cycle (both sexual, and asexual), the reef is fully occupied. A predetermined fraction (equal to 0.1) of the corals in the reef with the lowest health are selected, and eliminated to free up space for the next iteration of larvae to occupy. This process is repeated for 500 number of iterations, consequently, the healthiest coral is selected as the optimum solution.[20] The computational parameters used for CRO are tabulated in Table 2.

### 3.3 Bald Eagle Search

BES is an advanced, and recent swarm-evolutionary optimization algorithm that is motivated by the hunting technique used by bald eagles to prey on fishes. The BES algorithm is split into three stages: (i) identifying, and selecting the best area to search in the solution space, (ii) search in different directions within a spiral space for faster convergence towards optimal solution, and (iii) the eagles swoop towards the optimum solution, and each eagles takes a different shape while moving towards the optimum solution.

A population size of 20 is initialized to search the solution space. After each iteration the positional value of the particle is evaluated, this is done for each particle in the population. Initially an area is selected for searching around the best solution, and then the area is evaluated though a spiral movement. The solution moves from the next point to the central point, and is determined by a predefined number which is 10 in this paper. Further the next position is evaluated using a swooping movement. The number of search cycles is initialized to 1.5, and the movement intensity of each particle towards the best, and center points is initialized to 2.

The search space is randomly selected based on the previous search and is given by,

$$P_{new,i} = P_{best} + [\alpha \times r(P_{mean} - P_i)], \quad (5)$$

where, $r$ is a random value between 0 and 1 and is described in Table 2. $P_{new}$ and $P_{best}$ are the new and current search spaces. $P_{mean}$ signifies that these eagles have considered all the information from the previous points. The eagles search for solution in this space by moving in a spiral shape for faster computational time. In this stage, the eagle position is updated based on the following equations:

$$P_{i,new} = P_i + y(i) \times (P_i - P_{i+1})P_{best}] + [x(i) \times r(P_i - P_{mean})], \quad (6)$$

$$x(i) = \frac{xr(i)}{max|xr|}, \quad y(i) = \frac{yr(i)}{max|yr|}, \quad (7)$$

$$xr(i) = r(i) \times sin(\theta(i)), \quad yr(i) = r(i) \times cos(\theta(i)), \quad (8)$$

$$\theta(i) = \alpha \times \pi \times rand, \quad (9)$$

$$r(i) = \theta(i) \times R \times rand. \quad (10)$$

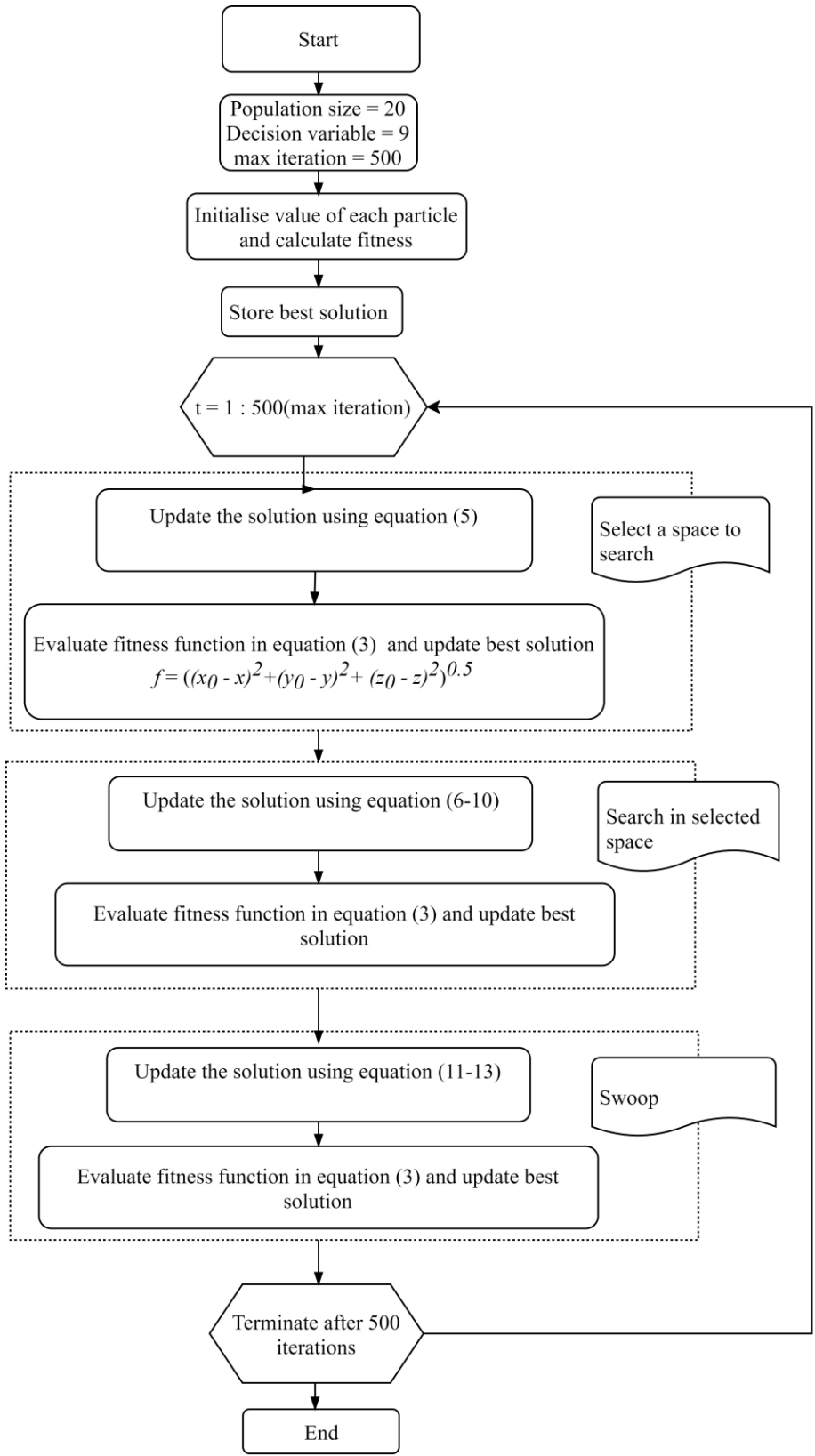

**Figure 3:** Flowchart of BES algorithm for solving IK problem of 9-DOF robot arm.

In the swooping stage the eagle moves towards the identified best solution in swooping manner given by equations:

$$P_{i,new} = rand * P_{best} + [x1(i) \times (P_i - c1 \times P_{mean})] + [y1(i) \times (P_i - c2 \times P_{best})] \quad (11)$$

$$x(i) = \frac{xr(i)}{max|xr|} \quad , \quad y(i) = \frac{yr(i)}{max|yr|}, \quad (12)$$

$$\theta(i) = \alpha \times \pi \times rand, \quad r(i) = \theta(i). \quad (13)$$

Once the search is completed, the point with the least value of fitness function is selected as the local best if it is lower than the previous best, and the process is repeated for 500 iterations. Finally, the best solution after 500 iterations is chosen as the global optimum.[16,24] The computational parameters used for BES are shown in Table 2. The working of the BES algorithm is given demonstrated in Figure 3.

### 3.4 Electromagnetic Field Optimization

EFO is a physics-based metaheuristic optimization algorithm that is inspired by the behavior of electromagnets of different polarity, and the nature's golden ratio. The polarity of this electro magnets can be changed by changing the direction of current in the coil. Electromagnets interact with other magnetic materials through attraction or repulsion. In EFO the ratio of attraction to repulsion is the golden ratio.

In EFO, each particle is represented by a group of electromagnets, and the number of electromagnets in each group signifies the number of variables. The particle that belongs to same electromagnets are of the same polarity. The electromagnet of same particles interacts with adjacent particle either with attractive or repulsive force. At first, like all MOA, a number of random particles are generated, and their fitness is evaluated. These particles are then sorted according to their fitness value. Later, the particles are divided into three groups as follows: (i) first group is the positive field that consist of particles of positive polarity (particles of highest fitness), (ii) second group is the negative field that consist of particles of negative polarity (particles of lowest fitness), and (iii) third group is the neutral field that consist of particles of mixed polarity, which creates a field of negative polarity that is very close to zero.

A random particle is generated, and its fitness is determined. According to the fitness of the new particle, it is sorted into one of the above three groups, and the least fit particle is discarded. Further, in correlation to the fitness of this new particle, the parameters of iteration of the next particle are altered to generate a more fit particle than the previous one. This process is continued until 500 iterations are complete, and the most positive group of electromagnets is selected as the optimal solution to the objective function.[17] The computational parameters used for EFO are shown in Table 2.

### 3.5 Multiverse Optimization

MVO Algorithm is another physics-based MOA that is inspired by the cosmological concepts of white holes, blackholes, and wormholes. MVO is used to perform exploration, exploitation, and local search, respectively. The mathematical modelling of the three mentioned concepts is the base of the MVO algorithm. The search process is categorized into exploration by the black hole, and white hole, and exploitation by the wormhole. Each solution in this method is analogous to a universe, and each variable in the solution is an object in the universe. The rules applied to the optimization are as follows, (a) higher inflation rate is inferred as higher the probability of having a white hole, and lower probability of black hole, (b) higher inflation rate increases the chances of sending objects from white holes, and simultaneously lowers the chances of receiving objects from blackholes, and (c) objects in any universe randomly moved to the best possible universe by using wormholes.

At first, each universe is tested against the objective function, and assigned an inflation rate. Then tunnels are established between multiple universes to exchange objects. The universe with higher inflation rate is assumed to have a white hole, whereas the one with lower inflation rate is assumed to have a blackhole. At the beginning of every iteration, the universes are sorted according to their inflation, and a random universe is picked to have a white hole. The objects are continuously exchanged in this manner to make each universe explore every part of the workspace in every iteration. In order to maintain the diversity among the universes, wormholes are placed in the mechanism to randomly transport objects between universes. As objects are being sent from universes of higher inflation rate to those of lower inflation rate, each universe is continuously being given a chance to improve its inflation rate. The above sequence of operations is run for 500 iterations, and at the end of all the iteration, all the universes will have similar values of inflation rate. Among these universes, the one with the highest inflation rate (fitness) is chosen as the global optimum solution to the objective function.[25] The computational parameters used for MVO are shown in Table 2.

Table 2: Value of parameters of different MOA.

| MOA | Parameter | Value |
|---|---|---|
| PSO | Accelerate constant towards local best, $c_1$ | 1.2 |
| | Accelerate constant towards global best, $c_2$ | 1.2 |
| | Inertia weight minimum | 0.4 |
| | Inertia weight maximum | 0.9 |
| CRO | Size of reef (n × m) | 1 × 20 |
| | The rate between free, and occupied during initialization | 0.4 |
| | Existing Corals rate | 0.9 |
| | Fraction of corals that duplicates itself, and tries to settle in a different part of the reef | 0.1 |
| | Fraction of the worse health corals in reef for which depredation will be applied | 0.1 |
| | Probability of depredation | 0.1 |
| | Number of attempts for a larva to set in the reef | 3 |
| BES | Determining the corner between point search in the central point, $a$ | 10 |
| | Determining the number of search cycles, $R$ | 1.5 |
| | Parameter for controlling the changes in position | 2 |
| | Movement intensity of bald eagles towards the best, and center points, $c_1, c_2$ | 2 |
| EFO | Probability of selecting electromagnets of generated particle from the positive field, PS-rate | 10 |
| | Probability of changing one electromagnet of generated particle with a random electromagnet, R-rate | 0.3 |
| | Portion of population having positive field | 0.1 |
| | Portion of population having negative field | 0.45 |
| MVO | Wormhole Existence Probability minimum, W_MIN | 0.2 |
| | Wormhole Existence Probability maximum, W_MAX | 1 |
| | Exploitation accuracy over the iterations, P | 6 |
| | Maximum iteration, L | 500 |
| NRO | Frequency of sinusoidal function | 0.05 |
| | Scaling factor, | 0.01 |
| | decay probability, $p_\beta$ | 0.1 |
| | Nuclear fission probability, $p_{Fi}$ | 0.75 |

## 3.6 Nuclear Reaction Optimization

NRO MOA is a search algorithm that is inspired by nuclear reactions viz. fission, and fusion. NRO assumes that the whole search space is a closed container, and all nuclei within the container will react with one another at all times. In this method, nuclear fission is used for exploitation, and nuclear fusion is used for exploration.

NRO uses a cycle of fission energy, and fusion electrons to find the most stable electron in the container, which is regarded as the optimal solution. Initially, a of set number of nuclei (sample solutions) is randomly selected in the workspace of the algorithm with respect to the upper, and lower bounds of the objective function. Each nucleus (position) may have many attributes (variables) associated with it. These attributes determine the binding energy (fitness) value of the nucleus to find the most stable nucleus.

In the nuclear fission phase, the nuclei are divided into odd, and even are formed. During an odd nucleus fission, a primary product, and a secondary product are formed. The secondary product that is more stable than primary product due to -decay. Besides, an even nucleus generally will not undergo the process of fission. On the other hand, In the nuclear fusion phase, two nuclei that are highly stable are fused to form a more stable nucleus. In NRO, nuclear fission process is carried out first, and then nuclear fusion. During multiple cycles of nuclear fission, and fusion of each nucleus in the container, the stability of the nucleus is increased. At the end of the 500 iterations, the most stable nucleus is assigned as the optimum solution to the objective function.[19,26]. The computational parameters used for NRO are tabulated in Table 2.

## 4. RESULTS AND DISCUSSION

For a pre-defined target point, *(x, y, z)* of the 9-DOF robot arm, the six MOA have been employed to find the final configuration. All the MOA were run for 200 generations with a population size of 10. After using fail, and error method, and different number of iterations, it was found that 500 iterations with a population size of 20 for all MOA. The position error, and the configuration of the nine joints for the pre-defined target point (1, 1, 7) is tabulated in Table 3. The configurations for all the MOA are shown in Figure 4. In Figure 5 the plot of the positional error versus iteration graphs of different MOA is depicted.

The PSO MOA was found to have low error of $10^{-11}$ cm for the target point (1, 1, 7) as seen in previous studies.[14] BES algorithm had shown the best performance among all MOA as it recorded 0 positional error for the target points (1, 1, 7) within the dexterous region. Physics based algorithms such as MVO, and NRO demonstrated $10^{-4}$ cm, and $10^{-3}$ cm positional error. EFO had $10^{-11}$ cm position error. In addition, EFO performed the best out of the all physics-based MOA. Evolutionary-based MOA, CRO had $10^{-4}$ cm position error.

**Table 3:** Configurations of robot arm for (1, 1, 7) for different MOA.

| Joint angle $\theta$ (°) | PSO | CRO | BES | EFO | MVO | NRO |
|---|---|---|---|---|---|---|
| $\theta_1$ | 225 | 45.16 | 45 | 45 | 44.48 | 225 |
| $\theta_2$ | 39.9 | -16.19 | 76.09 | 0.57 | 73.28 | 39.99 |
| $\theta_3$ | 2.35 | 76.39 | -51.2 | 10.17 | 11.17 | -13.8 |
| $\theta_4$ | 31.05 | -11.21 | -67.85 | 58.73 | -21.2 | 71.48 |
| $\theta_5$ | 59.3 | 90 | 79.26 | 68.95 | -29.1 | 47.26 |
| $\theta_6$ | -7.65 | 12.27 | 79.85 | -73.75 | -3.99 | -33.5 |
| $\theta_7$ | 62.77 | 15.81 | 5.314 | 21.85 | 147 | 30.66 |
| $\theta_8$ | -21.35 | -74.25 | -13.46 | 50.66 | 55.47 | 36.49 |
| $\theta_9$ | 9.741 | -65.67 | 72.22 | 65.05 | 112.4 | 14.28 |

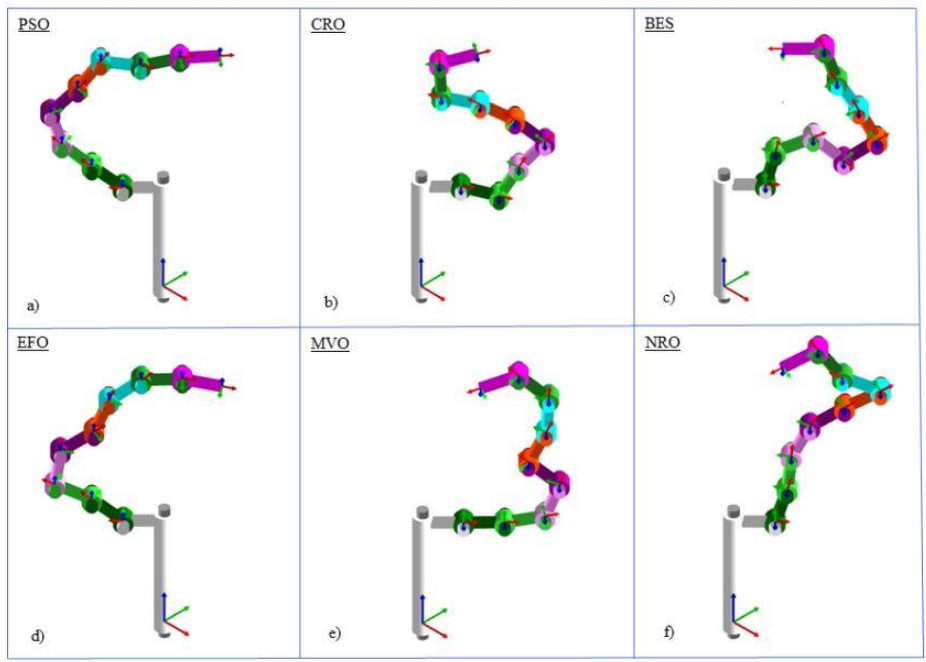

**Figure 4:** Visualization of robot configurations, clockwise from top left: a) PSO; b) CRO; c) BES; d) EFO; e) MVO, and f) NRO for the pre-defined point of robot arm using RoboAnalyzer.

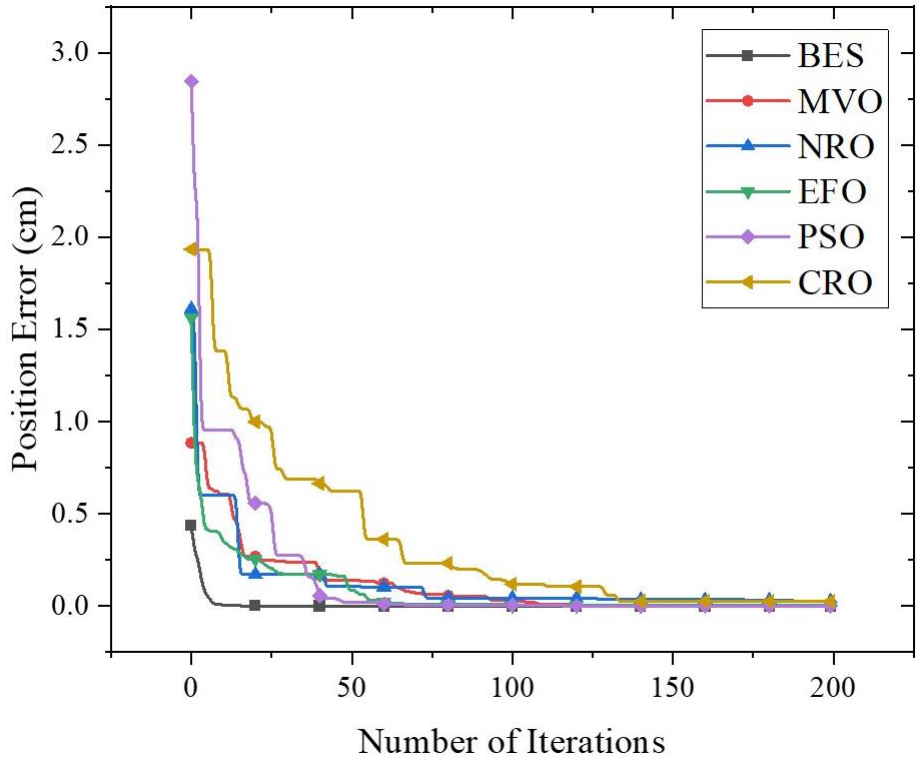

**Figure 5:** Comparison of the position error vs generation of the various metaheuristic algorithm for the point.

Further, ten random points common for all MOA with the base of robot situated at coordinate (0, 0, 0) shown in Figure 6 were considered in the dexterous workspace of the robot. The positional error for these 10, and points was calculated for all the MOAs. PSO showed a low average positional error of $3.15 \times 10^{-7}$ cm. BES outperformed all the other algorithms, and showed zero positional error for every point in the dexterous workspace. Interestingly, with a repeatability of 100% (*i.e.,* zero positional error was obtained for all number of times) of the time, and hence performed the best compared to all the algorithms. EFO, MVO, and NRO recorded an average positional error of $1.1 \times 10^{-11}$ cm, $3.9 \times 10^{-5}$ cm, and $7.4 \times 10^{-3}$ cm respectively for the ten random coordinate points. CRO showed a high average positional error of $1.2 \times 10^{-1}$ cm. The average positional error, and computational time for different MOA for 10 random points table is shown in Table 4.

**Table 4:** Average positional error, and computational time for different MOA for 10 random positions.

| Points (x, y, z) | Positional Error | | | | | |
|---|---|---|---|---|---|---|
| | PSO ($\times 10^{-9}$ cm) | CRO ($\times 10^{-3}$ cm) | BES ($\times 10^{-17}$ cm) | MVO ($\times 10^{-6}$ cm) | EFO ($\times 10^{-14}$ cm) | NRO ($\times 10^{-3}$ cm) |
| (-0.06, -0.04, 3.29) | 9.5 | 1.93 | 0 | 7.24 | 0.00555 | 7 |
| (0.16, -2.05, -3.77) | 23.7 | 10.7 | 0 | 110 | 0.0305 | 7 |
| (-0.30, 0.92, -4.24) | 6.8 | 8.1 | 0 | 0.268 | 1100 | 4 |
| (6.22, 1.15, 3.73) | 3100 | 1150 | 0 | 215 | 0 | 12 |
| (0.05, -0.02, 2.21) | 3.49 | 4.24 | 0 | 0.0634 | 0.00173 | 7 |
| (0.18, 1.46, 1.50) | 0.0042 | 4.13 | 0 | 0.754 | 6190 | 12 |
| (2.41, -0.17, 1.84) | 0.000016 | 0.725 | 0 | 5.37 | 3690 | 9 |
| (-2.92, -1.57, -1.30) | 6.7 | 1.78 | 0 | 1.13 | 0 | 4 |
| (0.06, 0.80, 2.26) | 0.0062 | 1.19 | 0 | 0.952 | 0.000693 | 8 |
| (0.42, -0.11, 8.75) | 0.87 | 9.8 | 0 | 49 | 0.00556 | 6 |
| Average Error | 315 | 120 | 0 | 39 | 1100 | 7.4 |
| Average Time (s) | 72 | 55 | 263 | 63 | 93 | 248 |

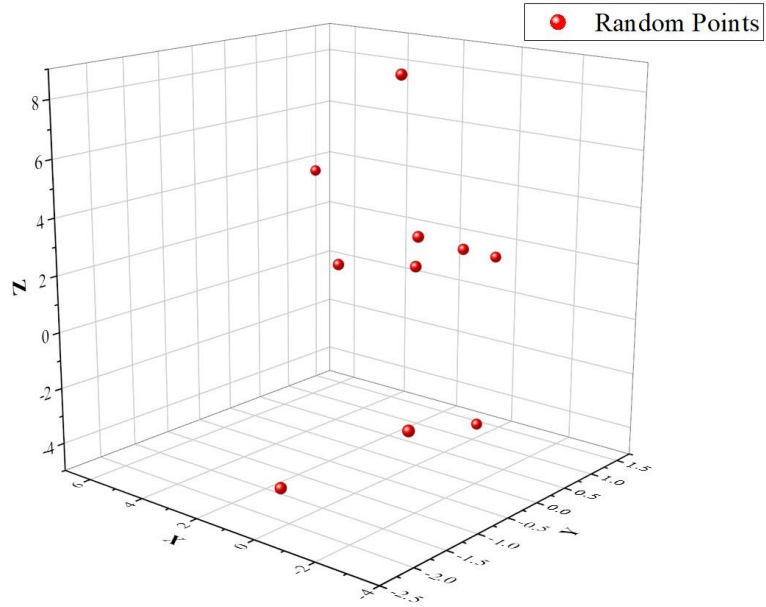

**Figure 6:** The 10 random points considered common for all MOA.

Bald Eagle Search outperforms all MOAs used as it takes advantage of synergy given by both swarm intelligence, and evolutionary algorithms. Like other MOA, BES initiates the search by initializing random particles at different positions on the workspace, and then evaluates the fitness of each particle. Later, the fittest particle is found from the fitness function value. In other commonly used swarm optimizers, the subsequent step is each particle has to move towards the general direction of the current best solution in a straight motion. If the distance covered by this motion results in an over-step of the location of the global minima, then the global minima will take more time to be found. This means that the points being searched are completely dependent on the initial random generation of particles, and the path taken by each particle. This kind of optimization technique will give rise to different solutions for the same objective function. On the other hand, BES does not move in a straight line towards the local current best position. It performs a unique swooping maneuver, in a swirl shape taking the distance between the current particle as the starting point, and the current best particle as the ending point, and searching each point in this swirl for a position that is better than the current best. If any better solutions are found within the swirl, between the starting, and ending points, the particle updates the new best position, eliminating the previous best. This process is carried out by each particle making a thorough search of the workspace. Finally, almost every point close to local minima is covered as swirls that intersect at many locations. This solution is an extremely precise, and accurate solution.

## 5. CONCLUSION

Six different MOA were implemented to the IK problem of 9-DOF robot arm, and an attempt was made to understand the accuracy, and precision of the MOA for the pre-defined target position. A common criterion of 500 iterations for a population size of 20 was set to different MOA. For the pre-defined target point the configuration of the robot was visualized using RoboAnalyzer. Additionally, 10 random points common for all MOA was considered, and their average positional error was compared. The position error versus generation was studied to understand the computation mechanism, and the time between different algorithms. It was observed BES achieved least positional error with least number of iterations due to synergetic effect of swarm intelligence, and evolutionary algorithm.

**Conflict of interest** There are no conflicts of interest to declare.

**Source of funding** This research received no specific funding from any funding agency.

**Author Contributions**

>Vineeth P. – Conceptualization, Data curation, Formal Analysis, Investigation, Methodology, Writing: review and editing.

>Guru Nanma P. – Formal Analysis, Data curation, Writing: review and editing.

>V. Sankar – Conceptualization, Data curation, Methodology, Writing original draft.

>B. Sachin Kumar – Investigation, Project administration, Resources, Supervision, Writing: review and editing.

**References**

1. Lopez-Franco C, Hernandez-Barragan J, Alanis AY, et al. A soft computing approach for inverse kinematics of robot manipulators. *Eng Appl Artif Intell* 2018; 74: 104–120.

2. Ayyıldız M, Çetinkaya K. Comparison of four different heuristic optimization algorithms for the inverse kinematics solution of a real 4-DOF serial robot manipulator. *Neural Comput Appl* 2016; 27: 825–836.

3. Dutra MS, Salcedo IL, Diaz LMP. New technique for inverse kinematics problem using Simulated Annealing. In: *Int. Conf. on Engineering Optimization*. Citeseer, 2008, 01–05.

4. Huang H-C, Chen C-P, Wang P-R. Particle swarm optimization for solving the inverse kinematics of 7-DOF robotic manipulators. In: *2012 IEEE international conference on systems, man, and cybernetics (SMC)*. IEEE, 2012, 3105–3110.

5. Kucuk S. Energy minimization for 3-RRR fully planar parallel manipulator using particle swarm optimization. *Mech Mach Theory* 2013; 62: 129–149.

6. Zhang Y, Wang S, Ji G. A comprehensive survey on particle swarm optimization algorithm and its applications. *Math Probl Eng* 2015; 931256: 1–38.

7. Adly MA, Abd-El-Hafiz SK. Inverse kinematics using single-and multi-objective particle swarm optimization. In: *2016 28th International Conference on Microelectronics (ICM)*. IEEE, 2016, 269–272.

8. Alkayyali M, Tutunji TA. PSO-based Algorithm for Inverse Kinematics Solution of Robotic Arm Manipulators. In: *2019 20th International Conference on Research and Education in Mechatronics (REM)*. IEEE, 2019, 1–6.

9. Rokbani N. IK-FA, a New Heuristic Inverse Kinematics Solver Using Firefly Algorithm. *Comput Intell Appl Model Control* 2015; 575: 1–28.

10. Pavan K K, Murali M J, Srikanth D. Generalized solution for inverse kinematics problem of a robot using hybrid genetic algorithms. *Int J Eng Technol* 2018; 7: 250–256.

11. Huang H-C, Xu SS-D, Wu CH. A hybrid swarm intelligence of artificial immune system tuned with Taguchi–genetic algorithm and its field-programmable gate array realization to optimal inverse kinematics for an articulated industrial robotic manipulator. *Adv Mech Eng* 2016; 8:1–10.

12. Köker R. A neuro-simulated annealing approach to the inverse kinematics solution of redundant robotic manipulators. *Eng Comput* 2013; 29: 507–515.

13. Köker R, Çakar T. A neuro-genetic-simulated annealing approach to the inverse kinematics solution of robots: a simulation based study. *Eng Comput* 2016; 32: 553–565.

14. Dereli S, Köker R. A meta-heuristic proposal for inverse kinematics solution of 7-DOF serial robotic manipulator: quantum behaved particle swarm algorithm. *Artif Intell Rev* 2020; 53: 949–964.

15. Zhang L, Xiao N. A novel artificial bee colony algorithm for inverse kinematics calculation of 7-DOF serial manipulators. *Soft Comput* 2019; 23: 3269–3277.

16. Alsattar H, Zaidan A, Bahaa B. Novel meta-heuristic bald eagle search optimisation algorithm. *Artif Intell Rev* 2020; 53: 2237–2264.

17. Abedinpourshotorban H, Shamsuddin SM, Beheshti Z, et al. Electromagnetic field optimization: A physics-inspired metaheuristic optimization algorithm. *Swarm Evol Comput* 2016; 26: 8–22.

18. Jalali SMJ, Khosravi A, Kebria PM, et al. Autonomous robot navigation system using the evolutionary multi-verse optimizer algorithm. In: *2019 IEEE International Conference on Systems, Man and Cybernetics (SMC)*. IEEE 2019; 1221–1226.

19. Wei Z, Huang C, Wang X, et al. Nuclear Reaction Optimization: A Novel and Powerful Physics-Based Algorithm for Global Optimization. *IEEE Access* 2019; 7: 66084–66109.

20. Salcedo-Sanz S, Ser JD, Landa-Torres I, et al. The Coral Reefs Optimization Algorithm: A Novel Metaheuristic for Efficiently Solving Optimization Problems. *Sci World J* 2014; 739768: 1–15.

21. Hayat AA, Sadanand RO, Saha SK. Robot manipulation through inverse kinematics. In: *Proceedings of the 2015 Conference on Advances In Robotics*. 2015, 48: 1–6.

22. Hayat A, Chittawadigi R, Udai A, et al. *Identification of Denavit-Hartenberg Parameters of an Industrial Robot*. 2013.

23. Rokbani N, Alimi AdelM. Inverse Kinematics Using Particle Swarm Optimization, A Statistical Analysis. *Procedia Eng* 2013; 64: 1602–1611.

24. Ramadan A, Kamel S, Hosny M, et al. An Improved Bald Eagle Search Algorithm for Parameter Estimation of Different Photovoltaic Models. *Processes* 2021; 9: 1127.

25. Mirjalili S, Mirjalili S, Hatamlou A. Multi-Verse Optimizer: a nature-inspired algorithm for global optimization. *Neural Comput Appl* 2016; 27: 495–513.

26. Ahmed AN, Van Lam T, Hung ND, et al. A comprehensive comparison of recent developed meta-heuristic algorithms for streamflow time series forecasting problem. *Appl Soft Comput* 2021; 105: 107282.